\documentclass[runningheads]{llncs}
\usepackage[T1]{fontenc}
\usepackage{graphicx}
\usepackage{ upgreek }
\usepackage{amsmath}
\usepackage{amsfonts}
\usepackage{amsmath, amssymb}
\usepackage{graphicx}
\usepackage{hyperref}

\usepackage{xspace}
\usepackage{float}
\usepackage{wrapfig}
\usepackage{makecell}
\usepackage{multirow}
\usepackage{graphicx}
\usepackage[table,xcdraw]{xcolor}
\usepackage{hyperref}

\usepackage{amsmath}
\usepackage{ dsfont }

\newcommand{\x}{\mathbf{x}}

\newcommand{\z}{\mathbf{z}}

\newcommand{\bphi}{{\boldsymbol{\phi}}}

\newcommand{\model}{LEDA}

\newcommand{\bv}{\mathbf{v}}
\newcommand{\bu}{\mathbf{u}}
\newcommand{\Diff}[0]{\mathrm{Diff}}

\begin{document}

\title{LEDA: Log-Euclidean Diffeomorphism Autoencoder for Efficient Statistical Analysis of Diffeomorphisms}

\titlerunning{\model}

\author{Krithika Iyer\inst{1, 2, 4} \and
Shireen Elhabian\inst{1,2} \and
Sarang Joshi\inst{1,3} }
%index{Iyer, Krithika} 
%index{Elhabian, Shireen} 
%index{Joshi, Sarang} 
%
\authorrunning{Iyer et al.}

\institute{Scientific Computing and Imaging Institute, University of Utah, UT, USA \and
Kahlert School of Computing, University of Utah, UT, USA \and Biomedical Engineering Department, University of Utah, UT, USA \and Corresponding Author \\
\email{\{krithika.iyer@ ,shireen@sci, sarang.joshi@\}.utah.edu}}

% \institute{Anonymous}
%
\maketitle              

\begin{abstract}
Image registration is a core task in computational anatomy that establishes correspondences between images. Invertible deformable registration, which computes a deformation field and handles complex, non-linear transformation, is essential for tracking anatomical variations, especially in neuroimaging applications where inter-subject differences and longitudinal changes are key.
Analyzing the deformation fields is challenging due to their non-linearity, limiting statistical analysis. However, traditional approaches for analyzing deformation fields are computationally expensive, sensitive to initialization, and prone to numerical errors, especially when the deformation is far from the identity.
To address these limitations, we propose the Log-Euclidean Diffeomorphism Autoencoder (LEDA), an innovative framework designed to compute the principal logarithm of deformation fields by efficiently predicting consecutive square roots. \model~operates within a linearized latent space that adheres to the diffeomorphisms group action laws, enhancing our model's robustness and applicability. We also introduce a loss function to enforce inverse consistency, ensuring accurate latent representations of deformation fields. 
Extensive experiments with the OASIS-1 dataset demonstrate the effectiveness of \model~in accurately modeling and analyzing complex non-linear deformations while maintaining inverse consistency. Additionally, we evaluate its ability to capture and incorporate clinical variables, enhancing its relevance for clinical applications. 

\end{abstract}

\section{Introduction}
%#############################################################################################
% Computational anatomy and its importance
\sloppy
The link between the form and function of anatomies is critical in understanding morphological variations to effectively diagnose diseases, plan procedures, and establish treatment methods \cite{asghar2024investigating}. Historically, observational studies were used to analyze anatomical variations \cite{alraddadi2021literature}. However, increasing accessibility of medical imaging technologies led to the availability of high-resolution in-vivo functional and structural imaging, providing a deeper understanding of organs. Consequently, computational anatomy has emerged as a critical tool to model, analyze, and quantify the variability of anatomical structures across individuals or populations \cite{ambellan2019statistical}. Image registration is a fundamental component of computational anatomy that introduces voxel-level/spatial correspondences. The voxel-level spatial correspondences transform a large dataset of images onto a standard coordinate frame to facilitate detailed morphological analysis. Image registration is widely used in applications such as neuroimaging studies, adaptive radiotherapy planning, and the development of population-specific atlases \cite{suganyadevi2022review,zachiu2020anatomically,binte2020spatiotemporal}. 

% Types of transformation 
\sloppy
There are two types of image transformation: rigid and non-rigid (or deformable) registration \cite{deng2022survey,yaniv2008rigid}. Images are aligned using simple translations and rotations in rigid image registration, which does not account for interior deformations. This method benefits solid structures like bones, where the relative spatial connections between image components stay constant. It is also effective in scenarios where no significant deformation is expected, such as intra-subject alignment of brain scans taken close in time, dental imaging, or preoperative and postoperative comparisons in orthopedics. 
In contrast, non-rigid registration accommodates complex, non-linear changes, which are essential for accurately aligning pre- and post-treatment images (e.g., in oncology), tracking progressive changes over time due to patient movement or disease progression, and constructing detailed anatomical atlases that reflect individual variability \cite{crum2004non}. 
Non-rigid methods can be broadly classified into parametric and non-parametric approaches \cite{hu2015tv}. Parametric methods use models such as B-splines or radial basis functions to represent the transformation in a structured and computationally efficient manner. Non-parametric methods estimate the deformation field directly from the data without assuming a specific functional form. This flexibility allows non-parametric methods to accommodate complex, non-linear anatomical variations, making them beneficial for highly deformable structures. 

% Diffeomorphic transforms
\sloppy
A smooth and invertible mapping that preserves the continuity and topology of anatomical structures is called a \textit{diffeomorphic} transform. These transformations ensure that no regions overlap or fold, making them ideal for capturing biologically plausible deformations. Methods for estimating deformation fields in computational anatomy span traditional mathematical approaches and modern deep learning techniques. 
Traditional methods, such as Large Deformation Diffeomorphic Metric Mapping (LDDMM) \cite{beg2005computing,hernandez2024insights,joshi2000landmark} and optical flow \cite{zhai2021optical}, focus on producing biologically plausible, smooth transformations. Techniques like Direct Deformation Estimation (DDE) \cite{boyle2019regularization} improve precision by directly computing deformation gradients. 
Deep learning models \cite{balakrishnan2019voxelmorph,chen2022transmorph,tian2023gradicon} leverage neural networks to efficiently learn complex deformation fields, enabling scalable and near real-time registration of large datasets. These advancements make deep learning approaches indispensable in clinical and research applications. 
However, ensuring inverse consistency—the symmetry and reversibility of transformations—is critical for reliable results, particularly in bidirectional or longitudinal studies. Loss functions intended to improve the consistency and robustness of learned transformations were incorporated into models such as GradICON \cite{tian2023gradicon} and ICON \cite{greer2021icon}. 

% Argument for diffeomorphisms as Lie groups and non-linear manifolds
\sloppy
Analyzing deformation fields is critical for detecting anatomical differences. While spatial correspondence is necessary, it cannot account for the complexities of these variations, necessitating the use of statistical tools to interpret the transformations \cite{avants2006geodesic,charon2013analysis}. However, conventional tools face significant challenges due to the complex, non-linear nature of diffeomorphic transforms \cite{pennec2015statistical}. The set of smooth, invertible transformations collectively forms the diffeomorphisms group of a manifold \(\mathcal{M}\), denoted as \(\Diff(\mathcal{M})\). This group is infinite-dimensional when \(\mathcal{M}\) is not zero-dimensional \(\mathrm{dim}(\mathcal{M})>0\) and simultaneously exhibits the structure of a Fréchet manifold and a Fréchet Lie group \cite{schmid2004infinite}. This duality arises because group operations—composition and inversion—are smooth, and the tangent space at the identity of this group corresponds to vector fields on \(\mathcal{M}\). The manifold structure of deformation fields introduces significant challenges. Unlike Euclidean spaces, where linear statistics such as means or Principal Component Analysis (PCA) are well-defined, these operations lack meaningful interpretation on curved manifolds like \(\Diff(\mathcal{M})\). Adding two deformation fields or directly averaging them in Euclidean terms fails to preserve the smoothness, invertibility, or geometric significance of the transformations. Such operations have no anatomical or mathematical relevance within the diffeomorphic framework.

\sloppy
Several methods such as Principal Geodesic Analysis (PGA) and its variants \cite{fletcher2004principal,zhang2013probabilistic,zhang2015bayesian}, Fréchet means \cite{le2000frechet}, and geodesic regression \cite{fletcher2011geodesic} have been developed to address the challenges of statistical estimation on manifolds. However, applying these methods to Fréchet Lie groups of diffeomorphisms requires significant adaptation due to the unique complexities of their structure. 
Adding a Riemannian metric to a Lie group to convert it into a Riemannian manifold is non-trivial, as not all Lie groups naturally possess such metrics. 
For example, a bi-invariant metric is a type of Riemannian metric that remains unchanged under left and right multiplication by group elements. They simplify computations, provide consistent geometric interpretations, and are particularly useful for analyzing symmetry. Compact Lie groups (e.g., $\mathrm{SO}(n)$, the group of rotations) have bi-invariant metrics; however, they are generally absent for non-compact Fréchet Lie groups such as $\Diff(\mathcal{M})$. This necessitates alternative meaningful metrics for non-compact groups to capture complex relationships between deformation fields in $\Diff(\mathcal{M})$ and ensure anatomically meaningful models.

% log-euclidean framework
\sloppy
The Log-Euclidean framework \cite{arsigny2006log} leverages the group structure of \(\Diff(\mathcal{M})\) instead of working directly on the non-linear manifold. The corresponding Lie algebra, represented by smooth vector fields, provides a linearized space for analyzing transformations.
This approach simplifies computations like geodesic distances and statistical averaging, which are challenging in the Riemannian setting due to non-linearity and curvature.  
The \textit{principal logarithm} maps group elements to their counterparts in the Lie algebra, enabling the representation of smooth, invertible transformations in a linear vector field space. This representation facilitates efficient computation of distances between transformations, offering an alternative to Riemannian approaches. However, optimization-based methods for estimating logarithms, such as the non-linear inverse scaling and square rooting algorithm \cite{arsigny2006log,kenney1989condition,higham2005scaling}, face challenges, including high computational cost, sensitivity to initialization, and susceptibility to noise. Moreover, these methods typically operate independently on each deformation field and do not consider the inverse consistency of transformations- a property crucial for ensuring reliable analyses in the Log-Euclidean framework \cite{miller2018computational}. 

\sloppy
These drawbacks underscore the need for more robust and efficient approaches to estimating principal logarithms. To address these limitations, we propose \textit{Log-Euclidean Diffeomorphism Autoencoder (LEDA)}, an innovative framework designed to compute the principal logarithm of deformation fields by efficiently predicting the consecutive square roots of deformation fields. The framework facilitates statistical analysis within a linearized latent space that respects the group action laws of the diffeomorphism group.
By appropriately mapping composition in the data space to scaling in the latent space, the framework enables the application of vector-space-based statistical methods, enhancing the robustness and applicability of statistical analysis for non-linear deformations.
Furthermore, we introduce a loss function to enforce inverse consistency constraints, ensuring the latent representations accurately capture the properties of the deformation fields.

\section{Related Work}
% early work with splines
The study of anatomical variability has led to the development of several computational methodologies for capturing complex, nonlinear changes inherent in biological structures. Diffeomorphism-based image registration is a method that helps capture nonlinear geometrical deformation in a population of images. Quantitatively comparing nonlinear registration algorithms necessitates computing global statistics about the deformation fields and is closely related to how diffeomorphisms are parameterized \cite{arsigny2006processing}. 

Marsland and Twining proposed using geodesic interpolating splines (GIS) \cite{camion2001geodesic} and polyharmonic clamped plate splines for low-dimensional representation of warps to enable statistical analysis. However, these methods are computationally expensive and unsuitable for complex invertible transformations frequently used in medical imaging \cite{marsland2004constructing}.
%
% Riemannian geometry frameworks
Pennec and Fillard developed a Riemannian geometry framework for statistical analysis on nonlinear spaces, particularly for anatomical structures \cite{pennec2015statistical}. Similarly, Principal Geodesic Analysis (PGA) \cite{fletcher2004principal} proposed by Fletcher et al. is designed explicitly for Riemannian manifolds and effectively utilizes geodesics to define principal components of underlying nonlinear manifolds. However, these approaches rely heavily on the choice of Riemannian metric, which may lack clear anatomical interpretation and entail significant computational overhead. 
%
% Time-varying velocity fields and momentum representation
Several studies have modeled diffeomorphisms as flows using time-varying velocity fields. Vaillant et al. proposed using the space of initial momentum as a linear representation of the nonlinear diffeomorphic shape space \cite{vaillant2004statistics}. While advantageous, it faces challenges such as high computational costs, limited interpretability of momentum representations, sensitivity to initialization, and difficulty handling large deformations beyond finite dimensional landmark matching.

% Deep Learning
More recently, Hinkle et al. proposed diffeomorphic autoencoders for Large Deformation Diffeomorphic Metric Mapping (LDDMM) specifically for atlas building \cite{hinkle2018diffeomorphic}, integrating momentum fields into diffeomorphisms through vector field flows governed by the Euler-Poincaré equation. Similarly, Bône et al. introduce diffeomorphic autoencoder \cite{bone2019learning} to simplify the shape analysis by using the vector momentum formulation of LDDMM. The approach leverages the EPDiff equation to ensure the transformations follow optimal paths between shapes. Although LDDMM supports statistical analysis of transformations by uniquely encoding shape by vectors normal to the outline of the template, they remain computationally expensive, limiting scalability to large datasets. 

In summary, while existing methods, including deep learning approaches, have advanced the modeling of anatomical variability using coordinate transformations, challenges remain in balancing scalability, computational efficiency, and interpretability. There is a continued need for robust frameworks that accurately capture complex transformations while ensuring efficiency and consistency.

% not talking about log-euclidean framework here as it has to be covered in detailed in the background section. 

\section{Background}
Here, we provide a brief overview of the theory and notations required for the Log-Euclidean framework. For a more detailed discussion, please refer to Vincent Arsigny \cite{arsigny2006processing} (Chapters 2 and 8).\\

\noindent \textbf{Log-Euclidean Framework: }A diffeomorphism \(\bphi: \mathbb{R}^D \to \mathbb{R}^D\) ($D=2$ for 2D images and $D=3$ for volumetric images) is a smooth, invertible mapping with a smooth inverse \(\bphi^{-1}\). 
The diffeomorphism \(\bphi \) maps every point \(\x\) in the original \(D\)-dimensional grid to a new location defined as \(\bphi(\x) = \x + \bu(\x)\), where \(\bu(\x)\) is the displacement field, while preserving topology and ensuring that no overlaps or folds occur.

Diffeomorphic transformations can be constructed by composing multiple small transformations. Specifically, \( \bphi \) can be expressed as \( \bphi(\x) = (\x + \epsilon \bv_{1}(\x)) \circ (\x + \epsilon \bv_{2}(\x)) \circ \cdots \circ (\x + \epsilon \bv_{n}(\x)), \) where each term \( (\x + \epsilon \bv_{i}(\x)) \) represents a small deformation controlled by the magnitude \(\epsilon \in \mathbb{R}^+\) and \( \bv_{i}(\x): \mathbb{R}^D \to \mathbb{R}^D \) is a smooth, bounded vector field. 
Each small deformation \((\x + \epsilon \bv_{i}(\x))\) is close to the identity when \(\epsilon\) is sufficiently small, ensuring smoothness and invertibility. The resulting full transformation \(\bphi\), remains diffeomorphic and smooth, though it may deviate from the identity depending on the cumulative effect of the vector fields \(\bv_{i}(\x)\).

The space of diffeomorphisms, \( \Diff(\mathcal{M}) \), forms a Lie group under composition, with the identity map serving as the group’s identity element; i.e., \(\bphi \circ \mathbf{id} = \mathbf{id} \circ \bphi = \bphi, \forall \bphi \in \Diff(\mathcal{M})\). 
The associated Lie algebra consists of smooth vector fields \( \bv(\x): \mathbb{R}^D \to \mathbb{R}^D\), which generate diffeomorphisms through their flows. These flows are described by one-parameter subgroups, \( \{\bphi_t\}_{t \in \mathbb{R}} \), a continuous family of diffeomorphisms satisfying the group property: \( \bphi_s \circ \bphi_t = \bphi_{s+t} \forall s, t \in \mathbb{R}, \) with \( \bphi_0 = \mathbf{id}\) as the identity map. 
The flows generated by these vector fields are governed by the ordinary differential equation (ODE) that describes how a point \(\x\) is transported by the vector field \(\bv\) over time \(t\) starting from the identity map at \(t = 0\):
\sloppy
\begin{equation}
    \frac{d\bphi_t(\x)}{dt} = \bv(\bphi_t(\x)), \quad \bphi_0(\x) = \x.
\end{equation}
The exponential map denoted as \( \exp(\bv) \), establishes a connection between the Lie algebra and the Lie group by generating a one-parameter subgroup \(\{\bphi_t\}_{t \in \mathbb{R}}\) from a vector field \( \bv \). 
%Conversely, the vector field \(\bv\), called the infinitesimal generator of the subgroup, can be recovered as:
% \begin{equation}
% \bv(\x) = \left.\frac{d}{dt}\right|_{t=0} \bphi_t(\x).
% \end{equation}
The logarithm map \(\log(\bphi)\) serves as the local inverse of the exponential map, allowing us to represent a diffeomorphism \(\bphi\) in terms of its generating vector field. Specifically, \(\log(\bphi) = \bv\) and \(\exp(\bv) = \bphi\).\\

\noindent \textbf{Logarithm Estimation: }
To compute the logarithm map \(\log(\bphi)\), a non-linear inverse scaling and square rooting algorithm has been proposed \cite{arsigny2006log}. This approach begins by selecting a scaling factor \(2^N\), which defines the number of successive square root operations needed to be computed. These operations iteratively transform \(\bphi\) into a version that is closer to the identity map \(\mathbf{id}\), where the logarithm map is more accurately approximated. 
Once \(N^{th}\) square root \(\bphi^{{-2}^N}\) is obtained, the logarithm of \(\bphi\) is estimated as:
\begin{equation}\label{log_map_estimation}
    \operatorname{log}(\bphi) = 2^N \operatorname{log}(\bphi^{{-2}^N}) \quad \text{where} \quad \operatorname{log}(\bphi^{{-2}^N}) = \bphi^{{-2}^N} - \mathbf{id} 
\end{equation}
However, this algorithm has several drawbacks. The repeated square root operations are computationally expensive, especially for high-dimensional data or when \(\bphi\) is far from the identity. Furthermore, the algorithm is sensitive to initialization, and the iterative process can accumulate numerical errors, reducing the accuracy of non-identity diffeomorphisms. 

%Finally, this approach is most reliable when \(\bphi\) is sufficiently close to the identity, as the accuracy of the logarithm diminishes with increasing distance from the identity.

% the diffeomorphisms are close to the identity, and its performance diminishes as the distance from the identity increases.

%The parameter \(\epsilon \in \mathbb{R}^+\) controls the magnitude of each deformation, and these small deformations are guaranteed to be diffeomorphic when \( \epsilon \) is sufficiently small, meaning the transformation remains close to the identity. 
%This ODE describes how a point \(\x\) is transported by the vector field \(\bv\) over time \(t\). Starting from the identity map at \(t = 0\) (\(\bphi_0(\x) = \x\)), the flow \(\bphi_t(\x)\) evolves continuously, tracing a smooth trajectory in the domain under the influence of \(\bv\).

\section{Methods}

\begin{figure}[htb]
    \centering
    \includegraphics[width=\linewidth]{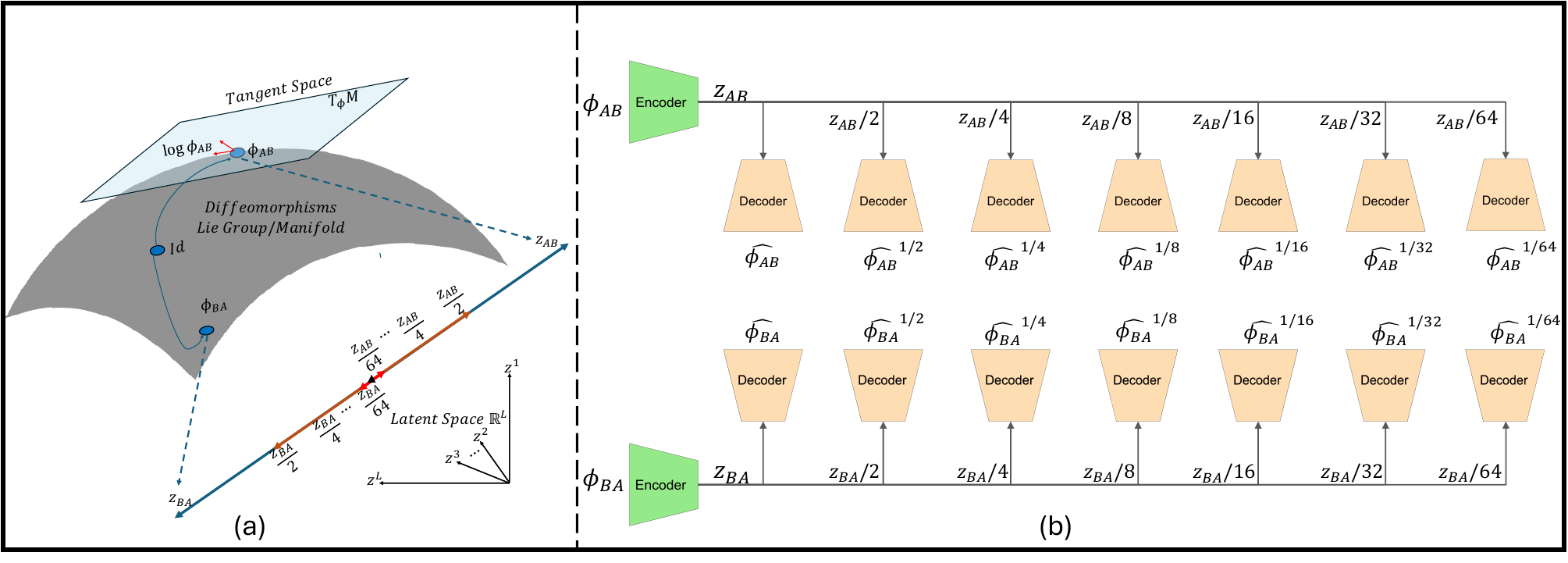}
    \vspace{-6mm}
    \caption{(a) \textbf{Mathematical framework for diffeomorphic transformations}: Shows the relationship between the Euclidean vector space (tangent space $T_{\phi}M$), the diffeomorphic group (Lie group/manifold), and the latent space ($\mathds{R}^L$). Here, A and B represent two coordinate spaces or imaging domains between which the deformations occurs. Transformations ($\bphi_{AB}, \bphi_{BA}$) and the logarithmic map ($\operatorname{log}(\bphi_{AB})$), alongside their projections into \model's latent space ($\z_{AB}, \z_{BA}$), demonstrating inverse consistency and the mapping of composition in the data space to scaling in the latent space. (b) \textbf{\model~architecture}}
    \label{model_architecture}
    \vspace{-6mm}
\end{figure}

\subsection{\model: Log-Euclidean Diffeomorphism Autoencoder}\label{LED_Method}

We introduce the Log-Euclidean Diffeomorphism Autoencoder (\model), a novel approach to efficiently estimate \(N\) successive square roots of the input deformation field. Such an iterative decomposition progressively reduces the deformation field closer to the identity, offering a computationally efficient and accurate approximation of the logarithm map, even for complex, high-dimensional deformations.
The \model~architecture (Figure~\ref{model_architecture}.b) includes an encoder \(f_{\gamma}(\bphi)\) that maps the input \(\bphi\) to a low-dimensional latent representation \(\z\in \mathbb{R}^L\). 
The decoder function \(g_{\theta}\) reconstructs the square roots of the deformation field, ensuring a consistent mapping between scaling in the latent space and composition in the deformation space. It predicts the \(n^{th}\) root of the deformation field using scaled versions of the latent representation \(\z\).
Mathematically, this can be expressed as:
\begin{equation}
    f_{\gamma}(\bphi) = \z \quad \quad g_{\theta}(\z/m) = \bphi^{1/m} ~~ \text{where~} m = 2^n ~~\forall n \in \{0, 1, \dots N\}
\end{equation}
The successive square roots estimation framework allows us to approximate the logarithm map using Equation~\ref{log_map_estimation}. 
The remaining \(N-1\) estimated roots contribute to establishing a direct relationship between latent space operations and deformation field compositions, enabling intuitive manipulation of complex spatial transformations. The \model~framework achieves its objectives through three core criteria incorporated into its design and loss function:

\vspace{0.05in}
\noindent \textbf{1. Faithful reconstruction:} The estimated roots must accurately reconstruct the original deformation field when composed a specified number of times i.e., if \(\bphi^{-m} \) is the predicted root at stage \(n\) where \(m = {2}^n\), then \( \bphi^{-m}\) composed \(m\) times should yield the original deformation field \(\bphi\).\\
\textbf{2. Inverse consistency}: The estimated roots should maintain inverse consistency, i.e., \(\bphi_{AB}(\bphi_{BA}(\x)) = \x\) for all \(\x\). Here, \( A \) and \( B \) represent two coordinate spaces  or imaging domains between which the deformations occur. To facilitate this, we model \model~as a \textbf{Siamese} \cite{koch2015siamese} autoencoder, that processes the forward \(\bphi_{AB}\) and inverse \(\bphi_{BA}\) fields simultaneously.\\
\textbf{3. Inverse consistency in latent space:} Representing the forward deformation \(\bphi_{AB} \) by \(\z_{AB}\) and the inverse field \(\bphi_{BA}\) by \(\z_{BA}\), inverse consistency in the latent space requires that \(\z_{AB} = -\z_{BA}\), ensuring the latent representations of a deformation field and its inverse are equal in magnitude but opposite in direction. 

These contributions establish a computationally efficient framework for analyzing and manipulating diffeomorphic transformations while preserving their key structural properties.

\subsection{\model: Architecture and Loss}

The \model~processes pairs of deformation fields \(({\bphi}_{AB,k}, {\bphi}_{BA,k})\) from the dataset 
\(\mathcal{D} = \{ ({\bphi}_{AB,k}, {\bphi}_{BA,k}): k = 1, 2, \dots, K\}\),
where \({\bphi}_{AB,k}\) represents a forward deformation field and \({\bphi}_{BA,k}\) is its corresponding inverse deformation field.
The Siamese nature of the \model~is illustrated in Figure~\ref{model_architecture}.b. It simultaneously processes one pair of deformation fields \(({\bphi}_{AB,k}, {\bphi}_{BA,k})\), using networks that share weights, as indicated by identical coloring in the figure. Although the decoder architecture in Figure~\ref{model_architecture}.b is unrolled to show the scaling of the latent space and corresponding root estimation, the implementation uses a single shared decoder network to perform these tasks. \model~framework is implemented using 2D convolutional layers and fully connected layers for the encoder and decoder. However, it can be straightforwardly extended to 3D.

To define the loss functions used by the model, we introduce the notation \(C_m(\bphi)\) to indicate the composition of \(\bphi\) with itself \(m\) times:
\begin{equation}
C_m(\bphi) = \underbrace{\bphi \circ \bphi \circ \dots \circ \bphi}_{m \text{ times}}
\end{equation} 

To ensure the model satisfies the criteria described in Section~\ref{LED_Method}, the training objective incorporates three loss terms:

\vspace{0.05in}
\noindent \textbf{Reconstruction loss:} Ensures accurate reconstruction of the original deformation field \(\bphi\) from its predicted roots. 
    \begin{equation}
        \mathcal{L}_{rec} = \sum^K_{k=1}\sum^N_{n=0, m = 2^n} \left\{ \left\|C_m(\widehat{\bphi}^{-m}_{AB,k}) - \bphi_{AB,k}\right\|^2 + \left\|C_m(\widehat{\bphi}_{BA,k}^{-m}) - \bphi_{BA,k}\right\|^2 \right\} 
    \end{equation}
\noindent where \(\bphi\) is the ground truth deformation field and \(\widehat{\bphi}^{-m}\) is the output of the decoder at the $n-$th stage, i.e., after scaling the latent representation by $1/2^n$.
    
\vspace{0.05in}
\noindent \textbf{Inverse consistency loss:} We use the approximate inverse consistency loss \cite{greer2021icon} proposed for image registration models to enforce inverse consistency for each estimated root.
    \begin{equation}
        \mathcal{L}_{inv} = \sum^K_{k=1}\sum^N_{n=0, m = 2^n}  \left\{ \left\| \widehat{\bphi}^{-m}_{AB,k} \circ \widehat{\bphi}^{-m}_{BA,k} - \mathbf{id} \right\|_2^2 + \left\| \widehat{\bphi}^{-m}_{BA,k} \circ \widehat{\bphi}^{-m}_{AB,k}- \mathbf{id} \right\|_2^2 \right\}
    \end{equation}

\vspace{0.05in}
\noindent \textbf{Latent inverse consistency loss:} Enforces latent inverse consistency using cosine similarity \(\Theta_k\) and magnitude constraints:
    \begin{equation}
        \mathcal{L}_{linv} =\sum^K_{k=1} \left\{ \frac{1 + \cos(\Theta_k)}{2} + \left\|\z_{AB,k} + \z_{BA,k}\right\|^2 \right\}
    \end{equation}
where \(\Theta_k\) representing the cosine similarity between \(\z_{AB,k}\) and \(\z_{BA,k}\).

The total loss function is given as: \(\mathcal{L} = \alpha_{rec}\mathcal{L}_{rec}  + \alpha_{inv}\mathcal{L}_{inv} + \alpha_{linv}\mathcal{L}_{linv}\) where \(\alpha\)'s represent the weight of each term. 
By explicitly linking latent space operations to deformation field compositions, \model~offers a robust and intuitive framework for manipulating complex spatial transformations. Its emphasis on inverse consistency ensures reliability, making it well-suited for applications requiring efficient diffeomorphic transformations.

\section{Experiments}
\begin{figure}[tb]
    \centering
    \includegraphics[width=\linewidth]{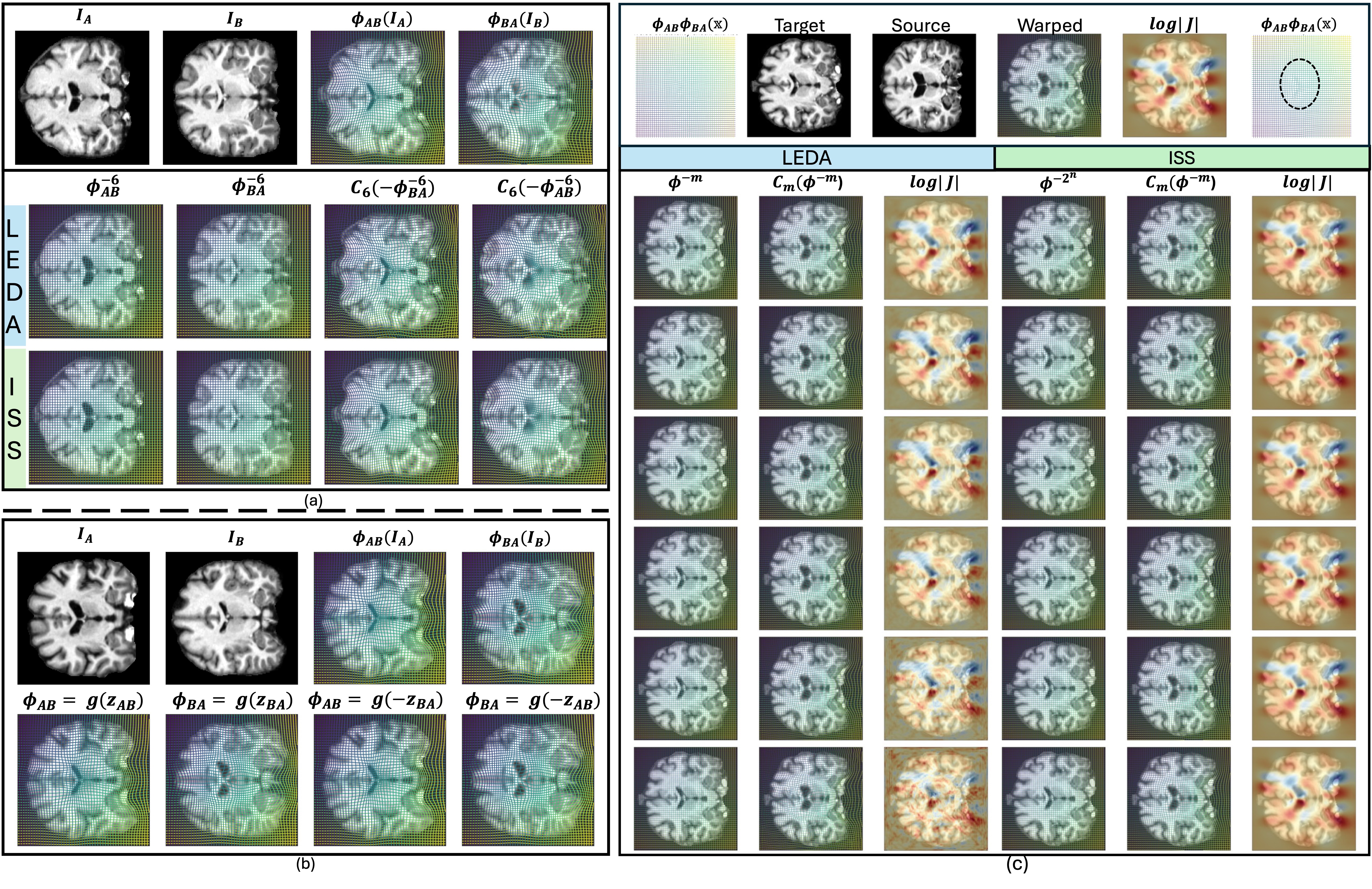}
    \vspace{-7mm}
    \caption{(a) Validation of Small Deformation Field Assumption (b) Validation of Latent Inverse Consistency (c) Comparison of Square Root Estimation\textbf{:} LEDA (left) and ISS (right)}
    \label{composition_check_and_negation}
\end{figure}
\begin{figure}[htb]
    \centering
    \includegraphics[width=\linewidth]{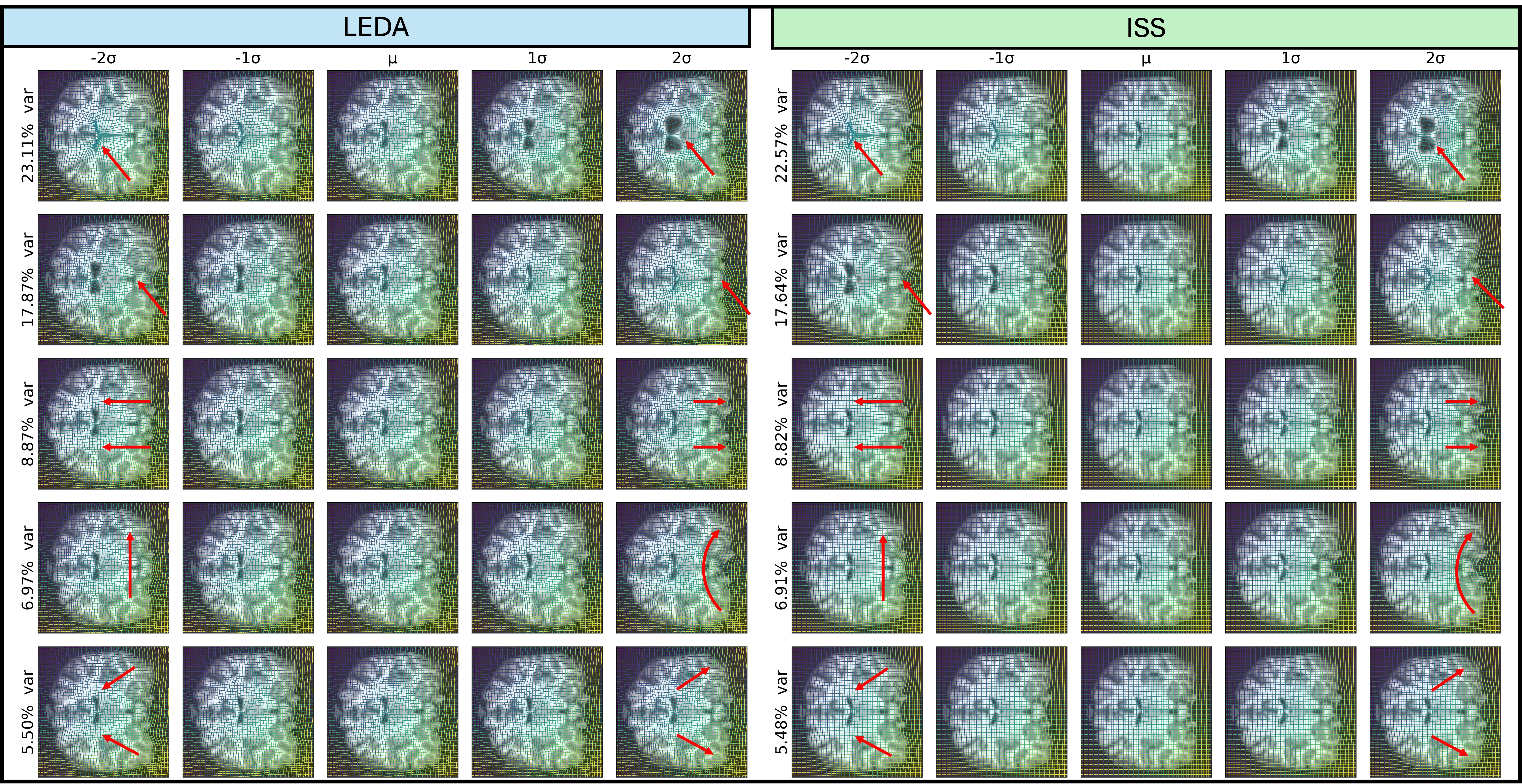}
    \vspace{-6mm}
    \caption{PCA Modes of Logarithm Maps \model~(left) and ISS (right): Variations can be associated with structural changes seen in AD, including ventricular expansion and hippocampal atrophy.}
    \label{data_pca}
\end{figure}
\begin{figure}[htb]
    \centering
    \includegraphics[width=\linewidth]{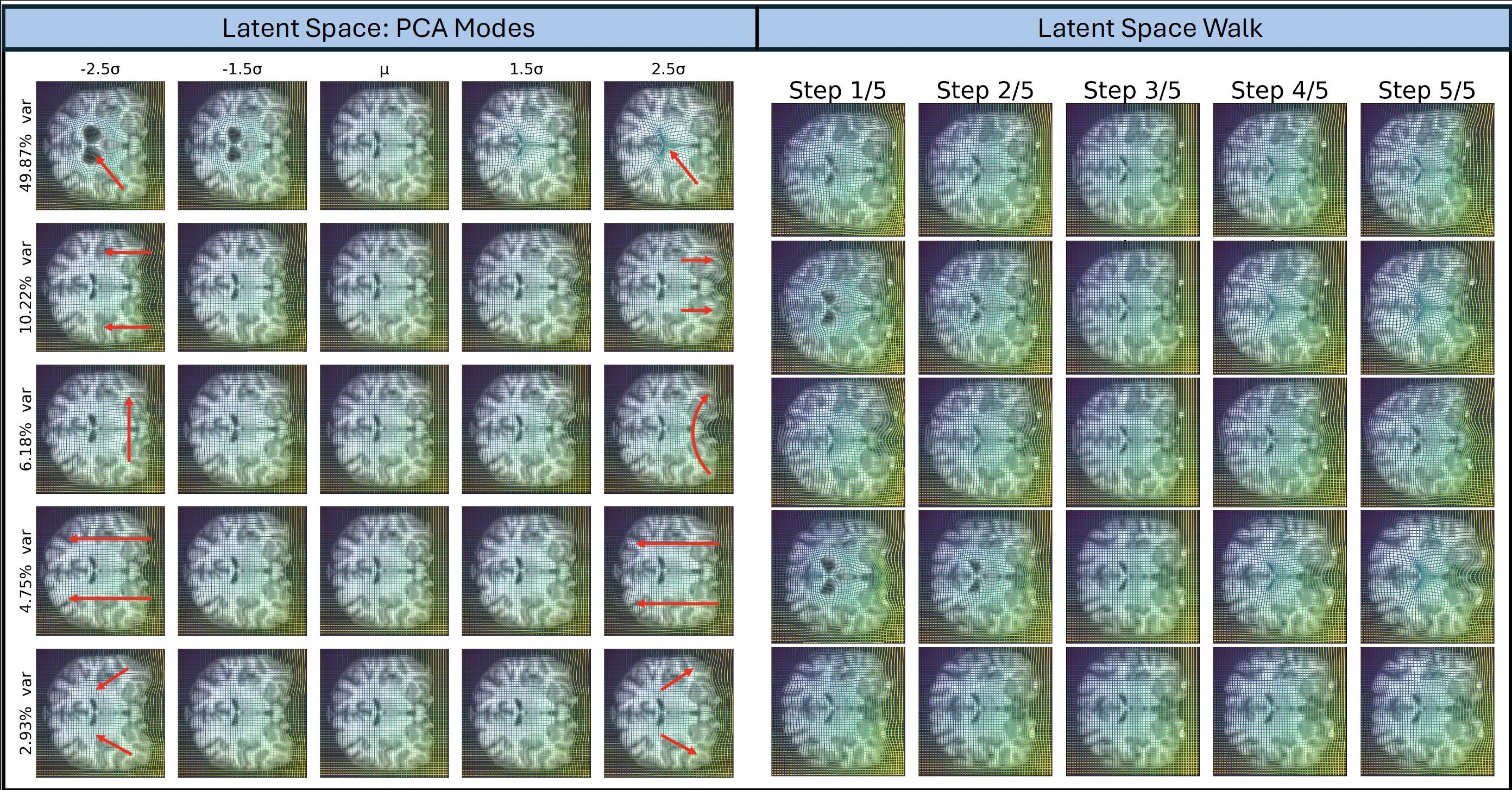}
    \vspace{-6mm}
    \caption{(a) PCA Modes of Latent Representations: Variations show structural changes associated with AD, including ventricular expansion and hippocampal atrophy. (b)\textbf{ }Latent Space Walk\textbf{:} Random walk demonstrates smooth, continuous transitions through anatomical variations.}
    \label{latent_pca_walk}
\end{figure}
\begin{figure}[htb]
    % \vspace{-2mm}
    \centering
    \includegraphics[width=\linewidth]{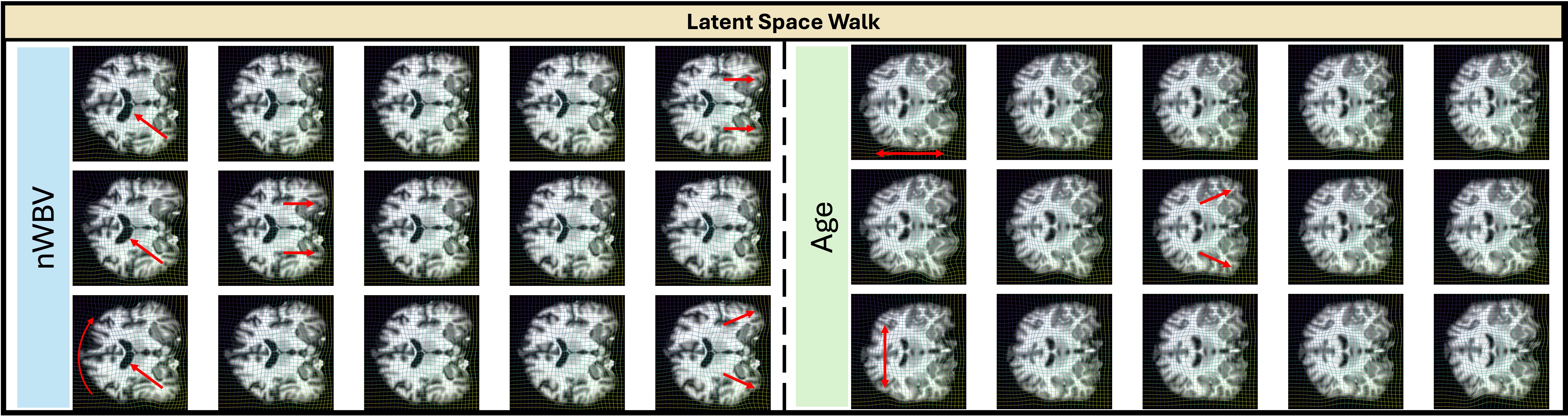}
    \vspace{-6mm}
    \caption{Latent Walk Along Latent Dimensions Predictive of Age and nWBV}
    \label{lr_indicative_dim_walk}
\end{figure}
\noindent \textbf{Dataset:} In this paper, we use the OASIS-1 dataset \cite{marcus2007open}, a widely-used neuroimaging resource containing 3D brain MRI scans. It includes multiple T1-weighted scans per subject, with 100 subjects over 60 diagnosed with very mild to moderate Alzheimer's disease (AD). We use 2D coronal slices of the scans \cite{hoopes2021hypermorph} and resize them to \(160 \times 160\). Deformation fields are generated by training a 2D GradICON model \cite{tian2023gradicon} on the 2D coronal slices dataset, producing 85,078 pairs of deformation fields from 413 2D images. 

\vspace{0.05in}
\noindent 
\textbf{Baseline:} 
We compare the logarithm maps and square root estimated from \model~with the non-linear inverse scaling and squaring (ISS) proposed by Arsigny \cite{arsigny2006log} to assess their ability to recover the original deformation. The authors use the closed from gradient updates, but we use PyTorch to estimate the gradients for square root estimation updates to simplify the process. 

\vspace{0.05in}
\noindent 
\textbf{Results:} Figure~\ref{composition_check_and_negation}.c shows a qualitative comparison of both methods. Notably, ISS is highly sensitive to initialization values, requiring a specific initialization of \(\frac{\bphi}{2}\). We tested multiple initializations and present results from the best-performing one. Both methods produce similar square root estimates, with higher roots approaching the identity transformation progressively (top to bottom). The log-determinant of the Jacobian reveals local deformations, where red indicates contraction and blue indicates expansion, and both methods show consistent patterns. However, ISS struggles with inverse consistency, as evidenced by the grid plots in the first row. While \model~fully recovers the identity grid, ISS leaves residual deformations, highlighting \model's superior robustness in maintaining inverse consistency. Moreover, estimating the square roots using ISS for a single deformation field takes \textbf{2.3 seconds} (given the PyTorch implementation), whereas performing inference via the trained \model~to estimate all the roots for a pair of deformation fields takes \textbf{only 0.02 seconds}. This substantial difference in computation time becomes even more pronounced when dealing with high-resolution deformation fields, underscoring the scalability and efficiency of the proposed framework.

We further verify the estimated logarithm maps to test their consistency with the small deformation field assumption, which states that forward displacement fields \(u_{AB}(\x)\) should approximate the negation of inverse displacement fields \(u_{BA}(\x)\). Using the \(2^6\)th root, we negate the forward displacement field \(\bphi^{-2^{6}}(\x)\), compose it appropriately, and recover the inverse field. Similarly, negating the inverse transform and composing it recovers the forward field. Figure~\ref{composition_check_and_negation}.a demonstrates that roots estimated by both methods satisfy this assumption, validating the accuracy of the logarithm maps and ensuring expected small deformation model behavior.

To assess inverse consistency within the latent space of \model, we negate the latent representation of forward and inverse displacement fields and decode them using \model.~Figure~\ref{composition_check_and_negation}.b shows that the \model~accurately decodes the negated latent representation into the corresponding inverse field, confirming inverse consistency. This validation is crucial as it demonstrates the \model's ability to represent and preserve relationships between forward and inverse transformations accurately.

We performed PCA on the logarithm maps of deformation fields predicted by both methods to analyze dominant modes of anatomical variation. Figure~\ref{data_pca} displays the identified modes aligning with known neuroanatomical changes in AD. The first mode (\(\sim23\%\) variance) captures large-scale atrophy patterns, including ventricular expansion and surrounding tissue reduction, particularly in the medial temporal lobe, consistent with AD pathology \cite{apostolova2012hippocampal,nestor2008ventricular}. The second mode (\(\sim18\%\)) highlights hippocampal atrophy and adjacent gray matter loss, reflecting early-to-moderate AD stages \cite{apostolova2012hippocampal}. This mode exhibits slight asymmetry, suggesting individual disease progression differences. The third mode (\(\sim9\%\)) reveals bilateral ventricular expansion with cortical thinning, while subsequent modes (\(\sim6-5\%\)) capture localized cortical and subcortical changes, such as finer tissue loss.

PCA was also conducted on the latent space of the \model,~with the modes of variation shown in Figure~\ref{latent_pca_walk}.a. The latent space modes closely align with the logarithm map PCA results, capturing clinically consistent changes such as ventricular expansion, hippocampus atrophy in early modes, and localized cortical thinning \cite{bakkour2009cortical}. Figure~\ref{latent_pca_walk}.b illustrates progressive interpolations in the latent space, visualized over five steps. These smooth and structured transitions effectively preserve anatomical coherence while capturing key variations, confirming the \ model's ability to generate realistic deformations and interpolate between anatomical states. 

Moreover, low-dimensional latent space representation is crucial for capturing essential features while reducing computational complexity, enabling more efficient analysis of high-dimensional neuroimaging data. We employed a linear regression framework to analyze deformation fields based on their latent presentations. The process involves (1) selecting a reference image $B$ of a healthy young adult, (2) computing deformation fields \(\bphi_{AB}\) for all samples $A$ excluding $B$, and (3) fitting linear regression models using the latent representations as independent variables and clinical variables from the OASIS dataset as dependent variables. This approach aims to uncover relationships between structural changes and clinical characteristics. The clinical variables included are age, Normalized Whole Brain Volume (nWBV), Mini-Mental State Examination (MMSE), Clinical Dementia Rating (CDR), estimated Total Intracranial Volume (eTIV), and Atlas Scaling Factor (ASF).
Our analysis indicated that the latent dimensions were most predictive of age and nWBV, with r-scores of 0.71 and 0.78, respectively. Figure~\ref{lr_indicative_dim_walk} illustrates the latent walk along the top three directions predictive of age and nWBV based on linear regression coefficients. Changes associated with age reflect overall brain shape and size, with minimal lateral ventricular expansion, consistent with the literature on age-related brain changes \cite{fujita2023characterization,curra2019ventricular}. In contrast, dimensions indicative of nWBV show significant alterations in ventricle shape and size, aligning with findings that decreased brain volume correlates with ventricular expansion in neurodegenerative conditions \cite{apostolova2013ventricular}.

\section{Conclusion}
In this work, we address the challenges of analyzing non-linear deformation fields in image registration by introducing a novel framework called Log-Euclidean Diffeomorphism Autoencoder (LEDA), designed to compute the principal logarithm of deformation fields by efficiently predicting the consecutive square roots of deformation fields. 
Extensive evaluations show LEDA's effectiveness in estimating logarithm maps that capture clinically relevant anatomical variations. LEDA's latent space can robustly link deformation fields to clinical variables, offering valuable insights into disease progression. With its efficiency and accuracy, LEDA opens avenues for efficient analysis of deformation fields, enabling more precise neuroimaging and medical applications.

\section*{Acknowledgement}
This work was supported by the National Institutes of
Health under grant numbers NIH-R01CA259686 (Sarang Joshi) and NIH-R01DE032366 (Shireen Elhabian).  

% \clearpage
\bibliographystyle{splncs04}
\bibliography{paper0181}

\begin{thebibliography}{10}
\providecommand{\url}[1]{\texttt{#1}}
\providecommand{\urlprefix}{URL }
\providecommand{\doi}[1]{https://doi.org/#1}

\bibitem{alraddadi2021literature}
Alraddadi, A.: Literature review of anatomical variations: clinical significance, identification approach, and teaching strategies. Cureus  \textbf{13}(4) (2021)

\bibitem{ambellan2019statistical}
Ambellan, F., Lamecker, H., von Tycowicz, C., Zachow, S.: Statistical shape models: understanding and mastering variation in anatomy. Springer (2019)

\bibitem{apostolova2013ventricular}
Apostolova, L.G., Babakchanian, S., Hwang, K.S., Green, A.E., Zlatev, D., Chou, Y.Y., DeCarli, C., Jack~Jr, C.R., Petersen, R.C., Aisen, P.S., et~al.: Ventricular enlargement and its clinical correlates in the imaging cohort from the adcs mci donepezil/vitamin e study. Alzheimer Disease \& Associated Disorders  \textbf{27}(2),  174--181 (2013)

\bibitem{apostolova2012hippocampal}
Apostolova, L.G., Green, A.E., Babakchanian, S., Hwang, K.S., Chou, Y.Y., Toga, A.W., Thompson, P.M.: Hippocampal atrophy and ventricular enlargement in normal aging, mild cognitive impairment (mci), and alzheimer disease. Alzheimer Disease \& Associated Disorders  \textbf{26}(1),  17--27 (2012)

\bibitem{arsigny2006processing}
Arsigny, V.: Processing data in lie groups: an algebraic approach. Ph.D. thesis (2006)

\bibitem{arsigny2006log}
Arsigny, V., Commowick, O., Pennec, X., Ayache, N.: A log-euclidean framework for statistics on diffeomorphisms. In: Medical Image Computing and Computer-Assisted Intervention--MICCAI 2006: 9th International Conference, Copenhagen, Denmark, October 1-6, 2006. Proceedings, Part I 9. pp. 924--931. Springer (2006)

\bibitem{asghar2024investigating}
Asghar, A., Patra, A., Naaz, S., Kumar, R., Babu, C.R., Singh, B.: Investigating the integration of anatomical variabilities into medical education as a potential strategy for mitigating surgical errors. Journal of the Anatomical Society of India  \textbf{73}(1),  70--81 (2024)

\bibitem{avants2006geodesic}
Avants, B.B., Epstein, C.L., Gee, J.C.: Geodesic image normalization and temporal parameterization in the space of diffeomorphisms. In: International Workshop on Medical Imaging and Virtual Reality. pp. 9--16. Springer (2006)

\bibitem{bakkour2009cortical}
Bakkour, A., Morris, J.C., Dickerson, B.C.: The cortical signature of prodromal ad: regional thinning predicts mild ad dementia. Neurology  \textbf{72}(12),  1048--1055 (2009)

\bibitem{balakrishnan2019voxelmorph}
Balakrishnan, G., Zhao, A., Sabuncu, M.R., Guttag, J., Dalca, A.V.: Voxelmorph: a learning framework for deformable medical image registration. IEEE transactions on medical imaging  \textbf{38}(8),  1788--1800 (2019)

\bibitem{beg2005computing}
Beg, M.F., Miller, M.I., Trouv{\'e}, A., Younes, L.: Computing large deformation metric mappings via geodesic flows of diffeomorphisms. International journal of computer vision  \textbf{61},  139--157 (2005)

\bibitem{binte2020spatiotemporal}
Binte~Alam, S., Nii, M., Shimizu, A., Kobashi, S.: Spatiotemporal statistical shape model for temporal shape change analysis of adult brain. Current Medical Imaging  \textbf{16}(5),  499--506 (2020)

\bibitem{bone2019learning}
B{\^o}ne, A., Louis, M., Colliot, O., Durrleman, S., Initiative, A.D.N.: Learning low-dimensional representations of shape data sets with diffeomorphic autoencoders. In: Information Processing in Medical Imaging: 26th International Conference, IPMI 2019, Hong Kong, China, June 2--7, 2019, Proceedings 26. pp. 195--207. Springer (2019)

\bibitem{boyle2019regularization}
Boyle, J.J., Soepriatna, A., Damen, F., Rowe, R.A., Pless, R.B., Kovacs, A., Goergen, C.J., Thomopoulos, S., Genin, G.M.: Regularization-free strain mapping in three dimensions, with application to cardiac ultrasound. Journal of biomechanical engineering  \textbf{141}(1),  011010 (2019)

\bibitem{camion2001geodesic}
Camion, V., Younes, L.: Geodesic interpolating splines. In: International workshop on energy minimization methods in computer vision and pattern recognition. pp. 513--527. Springer (2001)

\bibitem{charon2013analysis}
Charon, N.: Analysis of geometric and functional shapes with extensions of currents: applications to registration and atlas estimation. Ph.D. thesis, {\'E}cole normale sup{\'e}rieure de Cachan-ENS Cachan (2013)

\bibitem{chen2022transmorph}
Chen, J., Frey, E.C., He, Y., Segars, W.P., Li, Y., Du, Y.: Transmorph: Transformer for unsupervised medical image registration. Medical image analysis  \textbf{82},  102615 (2022)

\bibitem{crum2004non}
Crum, W.R., Hartkens, T., Hill, D.: Non-rigid image registration: theory and practice. The British journal of radiology  \textbf{77}(suppl\_2),  S140--S153 (2004)

\bibitem{curra2019ventricular}
Curr{\`a}, A., Pierelli, F., Gasbarrone, R., Mannarelli, D., Nofroni, I., Matone, V., Marinelli, L., Trompetto, C., Fattapposta, F., Missori, P.: The ventricular system enlarges abnormally in the seventies, earlier in men, and first in the frontal horn: a study based on more than 3,000 scans. Frontiers in Aging Neuroscience  \textbf{11}, ~294 (2019)

\bibitem{deng2022survey}
Deng, B., Yao, Y., Dyke, R.M., Zhang, J.: A survey of non-rigid 3d registration. In: Computer Graphics Forum. vol.~41, pp. 559--589. Wiley Online Library (2022)

\bibitem{fletcher2004principal}
Fletcher, P.T., Lu, C., Pizer, S.M., Joshi, S.: Principal geodesic analysis for the study of nonlinear statistics of shape. IEEE transactions on medical imaging  \textbf{23}(8),  995--1005 (2004)

\bibitem{fletcher2011geodesic}
Fletcher, T.: Geodesic regression on riemannian manifolds. In: Proceedings of the Third International Workshop on Mathematical Foundations of Computational Anatomy-Geometrical and Statistical Methods for Modelling Biological Shape Variability. pp. 75--86 (2011)

\bibitem{fujita2023characterization}
Fujita, S., Mori, S., Onda, K., Hanaoka, S., Nomura, Y., Nakao, T., Yoshikawa, T., Takao, H., Hayashi, N., Abe, O.: Characterization of brain volume changes in aging individuals with normal cognition using serial magnetic resonance imaging. JAMA Network Open  \textbf{6}(6),  e2318153--e2318153 (2023)

\bibitem{greer2021icon}
Greer, H., Kwitt, R., Vialard, F.X., Niethammer, M.: Icon: Learning regular maps through inverse consistency. In: Proceedings of the IEEE/CVF International Conference on Computer Vision. pp. 3396--3405 (2021)

\bibitem{hernandez2024insights}
Hernandez, M., Julvez, U.R.: Insights into traditional large deformation diffeomorphic metric mapping and unsupervised deep-learning for diffeomorphic registration and their evaluation. Computers in Biology and Medicine  \textbf{178},  108761 (2024)

\bibitem{higham2005scaling}
Higham, N.J.: The scaling and squaring method for the matrix exponential revisited. SIAM Journal on Matrix Analysis and Applications  \textbf{26}(4),  1179--1193 (2005)

\bibitem{hinkle2018diffeomorphic}
Hinkle, J., Womble, D., Yoon, H.J.: Diffeomorphic autoencoders for lddmm atlas building  (2018)

\bibitem{hoopes2021hypermorph}
Hoopes, A., Hoffmann, M., Fischl, B., Guttag, J., Dalca, A.V.: Hypermorph: Amortized hyperparameter learning for image registration. In: Information Processing in Medical Imaging: 27th International Conference, IPMI 2021, Virtual Event, June 28--June 30, 2021, Proceedings 27. pp. 3--17. Springer (2021)

\bibitem{hu2015tv}
Hu, W.R., Xie, Y., Li, L., Zhang, W.S.: A tv-l 1 based nonrigid image registration by coupling parametric and non-parametric transformation. International Journal of Automation and Computing  \textbf{12}(5),  467--481 (2015)

\bibitem{joshi2000landmark}
Joshi, S.C., Miller, M.I.: Landmark matching via large deformation diffeomorphisms. IEEE transactions on image processing  \textbf{9}(8),  1357--1370 (2000)

\bibitem{kenney1989condition}
Kenney, C., Laub, A.J.: Condition estimates for matrix functions. SIAM Journal on Matrix Analysis and Applications  \textbf{10}(2),  191--209 (1989)

\bibitem{koch2015siamese}
Koch, G., Zemel, R., Salakhutdinov, R., et~al.: Siamese neural networks for one-shot image recognition. In: ICML deep learning workshop. vol.~2, pp. 1--30. Lille (2015)

\bibitem{le2000frechet}
Le, H., Kume, A.: The fr{\'e}chet mean shape and the shape of the means. Advances in Applied Probability  \textbf{32}(1),  101--113 (2000)

\bibitem{marcus2007open}
Marcus, D.S., Wang, T.H., Parker, J., Csernansky, J.G., Morris, J.C., Buckner, R.L.: Open access series of imaging studies (oasis): cross-sectional mri data in young, middle aged, nondemented, and demented older adults. Journal of cognitive neuroscience  \textbf{19}(9),  1498--1507 (2007)

\bibitem{marsland2004constructing}
Marsland, S., Twining, C.J.: Constructing diffeomorphic representations for the groupwise analysis of nonrigid registrations of medical images. IEEE transactions on medical imaging  \textbf{23}(8),  1006--1020 (2004)

\bibitem{miller2018computational}
Miller, M.I., Arguill{\`e}re, S., Tward, D.J., Younes, L.: Computational anatomy and diffeomorphometry: A dynamical systems model of neuroanatomy in the soft condensed matter continuum. Wiley Interdisciplinary Reviews: Systems Biology and Medicine  \textbf{10}(6),  e1425 (2018)

\bibitem{nestor2008ventricular}
Nestor, S.M., Rupsingh, R., Borrie, M., Smith, M., Accomazzi, V., Wells, J.L., Fogarty, J., Bartha, R., Initiative, A.D.N.: Ventricular enlargement as a possible measure of alzheimer's disease progression validated using the alzheimer's disease neuroimaging initiative database. Brain  \textbf{131}(9),  2443--2454 (2008)

\bibitem{pennec2015statistical}
Pennec, X., Fillard, P.: Statistical computing on non-linear spaces for computational anatomy. Handbook of Biomedical Imaging: Methodologies and Clinical Research pp. 147--168 (2015)

\bibitem{schmid2004infinite}
Schmid, R.: Infinite dimentional lie groups with applications to mathematical physics  (2004)

\bibitem{suganyadevi2022review}
Suganyadevi, S., Seethalakshmi, V., Balasamy, K.: A review on deep learning in medical image analysis. International Journal of Multimedia Information Retrieval  \textbf{11}(1),  19--38 (2022)

\bibitem{tian2023gradicon}
Tian, L., Greer, H., Vialard, F.X., Kwitt, R., Est{\'e}par, R.S.J., Rushmore, R.J., Makris, N., Bouix, S., Niethammer, M.: Gradicon: Approximate diffeomorphisms via gradient inverse consistency. In: Proceedings of the IEEE/CVF Conference on Computer Vision and Pattern Recognition. pp. 18084--18094 (2023)

\bibitem{vaillant2004statistics}
Vaillant, M., Miller, M.I., Younes, L., Trouv{\'e}, A.: Statistics on diffeomorphisms via tangent space representations. NeuroImage  \textbf{23},  S161--S169 (2004)

\bibitem{yaniv2008rigid}
Yaniv, Z.: Rigid registration. Image-guided interventions: technology and applications pp. 159--192 (2008)

\bibitem{zachiu2020anatomically}
Zachiu, C., De~Senneville, B.D., Willigenburg, T., van Zyp, J.V., De~Boer, J., Raaymakers, B.W., Ries, M.: Anatomically-adaptive multi-modal image registration for image-guided external-beam radiotherapy. Physics in Medicine \& Biology  \textbf{65}(21),  215028 (2020)

\bibitem{zhai2021optical}
Zhai, M., Xiang, X., Lv, N., Kong, X.: Optical flow and scene flow estimation: A survey. Pattern Recognition  \textbf{114},  107861 (2021)

\bibitem{zhang2015bayesian}
Zhang, M., Fletcher, P.T.: Bayesian principal geodesic analysis for estimating intrinsic diffeomorphic image variability. Medical image analysis  \textbf{25}(1),  37--44 (2015)

\bibitem{zhang2013probabilistic}
Zhang, M., Fletcher, T.: Probabilistic principal geodesic analysis. Advances in neural information processing systems  \textbf{26} (2013)

\end{thebibliography}

\end{document}